# Cooperation between Top-Down and Bottom-Up Theorem Provers

**Dirk Fuchs**                                            DFUCHS@INFORMATIK.UNI-KL.DE
*Fachbereich Informatik, Universität Kaiserslautern*
*67663 Kaiserslautern, Germany*

**Marc Fuchs**                                            FUCHSM@INFORMATIK.TU-MUENCHEN.DE
*Fakultät für Informatik, TU München*
*80290 München, Germany*

## Abstract

Top-down and bottom-up theorem proving approaches each have specific advantages and disadvantages. Bottom-up provers profit from strong redundancy control but suffer from the lack of goal-orientation, whereas top-down provers are goal-oriented but often have weak calculi when their proof lengths are considered. In order to integrate both approaches, we try to achieve cooperation between a top-down and a bottom-up prover in two different ways: The first technique aims at supporting a bottom-up with a top-down prover. A top-down prover generates subgoal clauses, they are then processed by a bottom-up prover. The second technique deals with the use of bottom-up generated lemmas in a top-down prover. We apply our concept to the areas of model elimination and superposition. We discuss the ability of our techniques to shorten proofs as well as to reorder the search space in an appropriate manner. Furthermore, in order to identify subgoal clauses and lemmas which are actually relevant for the proof task, we develop methods for a relevancy-based filtering. Experiments with the provers SETHEO and SPASS performed in the problem library TPTP reveal the high potential of our cooperation approaches.

## 1. Introduction

Automated deduction is—at its lowest level—a search problem that spans huge search spaces. In the past many different calculi have been developed in order to cope with problems from the area of automated theorem proving. Essentially, for first-order theorem proving two main paradigms for calculi are in use: *Top-down calculi* like *model elimination* (ME, Loveland, 1968, 1978) attempt to recursively break down and transform a goal into subgoals that can finally be proven immediately with the axioms or with assumptions made during the proof. *Bottom-up calculi* like *superposition* (e.g., Bachmair & Ganzinger, 1994) go the other way by producing logic consequences from the input set until an obvious inconsistency is derived.

When comparing results of various provers (e.g., Sutcliffe & Suttner, 1997) it is obvious that provers based on different paradigms often behave quite differently. There are problems where bottom-up theorem provers perform considerably well, but top-down provers poorly, and vice versa. The main reason for this is that bottom-up provers often suffer from the lack of goal-orientation of their search, but profit from their strong redundancy control mechanisms. In contrast, top-down provers profit from their goal-orientation but suffer from insufficient redundancy control. This entails long proof lengths for many problems (e.g., Letz et al., 1994). Therefore, a topic that has come into the focus of research is the integration of both approaches. Specifically, cooperation between theorem provers (e.g., Conry et al., 1990; Schumann, 1994; Denzinger, 1995; Bonacina & Hsiang, 1995; Bonacina, 1996; Wolf & Fuchs, 1997; Fuchs, 1998b, 1998c) based on top-down and bottom-up principles appears to be promising because by exchanging information each approach can profit from the other. It is also possible to modify calculi or provers which work according to one paradigm so as to introduce aspects of the other paradigm into it. This, however, requires a lot of implementational effort to modify the provers, whereas our approach does not require changes of the provers but only changes of their input. We can hence employ arbitrary state-of-the-art provers.

Information that is well-suited for improving the performance of top-down provers are lemmas deduced by bottom-up provers. These lemmas are added to the input of a top-down prover and can help to shorten the proof length by immediately solving subgoals. Normally, the employed proof procedures can significantly profit from the proof length reduction obtained. The use of lemmas, however, also imports additional redundancy into the calculus. This means that an unbounded use of bottom-up generated lemmas without using techniques for choosing only *relevant* lemmas (i.e. lemmas which lead to a reduction of the search effort to enumerate a proof) is not sensible. In this article, in contrast to other approaches which generate lemmas dynamically during the proof run (Astrachan & Stickel, 1992; Astrachan & Loveland, 1997), we want to employ a bottom-up prover for generating lemmas in a preprocessing phase. After the generation of a pool of *lemma candidates* relevant lemmas are selected from this pool and the formula to be refuted is augmented by these bottom-up generated formulas.

The second main aspect that we consider is top-down/bottom-up integration by transferring information from a top-down prover to a bottom-up prover (e.g., Fuchs, 1998a). Our approach is to transfer top-down generated *subgoal clauses*—which essentially represent a transformation of an original goal clause into subgoals—to a bottom-up prover and to augment its input by these clauses. This introduces a goal-oriented component into a bottom-up prover which can enable it to solve proof problems considerably faster. However, as is the case with lemmas, an unbounded transfer of subgoal clauses is not sensible. Thus, we generate again subgoal clauses in a preprocessing phase and integrate only some of these clauses into the input set of a bottom-up prover. This necessitates techniques for *selecting relevant subgoal clauses*, i.e. techniques for selecting a set of subgoal clauses which can decrease the search effort of a bottom-up prover in order to find a proof.

In order to examine this kind of top-down/bottom-up integration we restrict ourselves to the bottom-up superposition calculus and the top-down connection tableau calculus. These calculi are very important since they are the basis for many high-performance theorem provers. For instance, the bottom-up provers SPASS (Weidenbach et al., 1996) and Gandalf (Tammet, 1997) that were most successful in recent proving competitions employ superposition and ordered paramodulation, respectively. The connection tableau calculus (or its restriction model elimination) is also the basis for very successful top-down provers, e.g., SETHEO (Moser et al., 1997) or METEOR (Astrachan & Loveland, 1991). In our opinion the concepts introduced for superposition and the connection tableau calculus can rather easily be transferred to other bottom-up and top-down calculi. Hence, the choice of these two calculi surely is justified.

The article is organized as follows. We start with a brief overview of superposition and model elimination (Section 2). Moreover, we discuss strengths and weaknesses of the calculi in more detail and introduce our approach for combining the strengths of both calculi. In Section 3 we address effects of the integration of ME subgoal clauses into the search state of a superposition-based prover. Furthermore, we describe two variants of a relevancy-based filtering of subgoal clauses. Section 4 deals with the use of bottom-up generated lemmas. We discuss in detail the ability of the produced clauses in order to help decrease proof lengths for refuting a given set of clauses as well as to reorder the search space in an appropriate manner. Based on this discussion, we introduce several relevancy measures. In Section 5, an experimental study conducted with the high-performance theorem provers SETHEO and SPASS reveals the potential of our techniques. We have chosen these systems in order to show that our concept can easily be integrated into existing systems and is even able to improve on the performance of very powerful provers. Finally, in Section 6 an overview of related approaches for top-down/bottom-up integration concludes the article.

## 2. A Framework for Coupling Top-Down and Bottom-Up Provers

In the following, we introduce typical representatives of top-down and bottom-up calculi and discuss their strengths and weaknesses in detail. After that, we sketch the basics of our methodology in order to combine these calculi.

### 2.1 Automated Theorem Proving with Superposition and Model Elimination

The general problem in first-order theorem proving is to show the inconsistency of a set $\mathcal{C}$ of clauses. A clause is a set of literals. As already discussed, theorem provers usually utilize either top-down or bottom-up calculi for accomplishing this task.

Typically, a bottom-up calculus contains several inference rules which can be applied to a set of clauses that constitute the search state. Generally, the inference rules can be divided into two classes: *Expansion* inference rules permit the generation of new clauses and *contraction* inference rules delete clauses or replace them by others. The most popular bottom-up calculus is the *resolution* calculus (Robinson, 1965). There, the expansion rules are resolution and factoring. The resolution calculus can be extended with contraction rules, e.g. the deletion of tautologies. If equality is involved in a problem it is sensible to employ the superposition calculus (Bachmair & Ganzinger, 1994), which extends resolution with specific rules suitable for handling equations. The expansion rules of the superposition calculus are superposition, equality resolution, and equality factoring. Again, additional contraction rules such as tautology deletion, subsumption, condensing, and rewriting can be employed. It is to be emphasized that for our study we employ the version of the superposition calculus introduced by Bachmair and Ganzinger (1994). Specifically this entails that factoring is only applied to positive literals.

A bottom-up theorem prover usually maintains a set $\mathcal{F}^P$ of so-called *potential* or *passive clauses* from which it selects and removes one clause $C$ at a time. This clause is put into the set $\mathcal{F}^A$ of *activated clauses*. Activated clauses are, unlike potential clauses, allowed to produce new clauses via the application of some inference rules. The inferred new clauses are put into $\mathcal{F}^P$. Initially, $\mathcal{F}^A = \emptyset$ and $\mathcal{F}^P = \mathcal{C}$. The indeterministic selection or *activation* *step* is realized by heuristic means. To this end, a heuristic $\mathcal{H}$ associates a natural number $\omega_C \in \mathbb{N}$ with each $C \in \mathcal{F}^P$, and the $C \in \mathcal{F}^P$ with the smallest weight $\omega_C$ is selected. An important property of heuristics is their *fairness*. A heuristic is called fair if it selects potential clauses in such a manner that no clause remains passive infinitely long. Usually the fairness of the used heuristic implies that the prover is complete, i.e. it can derive the empty clause when obtaining an inconsistent input set (provided the underlying calculus is complete).

The main strength of bottom-up calculi and provers is their strong redundancy control. On the one hand, many inferences which are definitely unnecessary in a proof, e.g. inferences involving tautologies, are omitted. On the other hand, contraction inference rules like subsumption avoid the repetition of expansion inferences involving the same (or more instantiated) clauses. A big disadvantage of bottom-up calculi is their lack of goal-orientation. Because certain inferences are favored over others due to the fixed search strategy and the heuristic weight of the clauses part of it, it might be the case that for a very long time only clauses which are not part of any proof are enumerated.

Model elimination is a typical top-down calculus which we shall introduce in the form of the *connection tableau calculus* ($CTC$) (Letz et al., 1994). In order to introduce $CTC$ we want to start with the basic (free variable) tableau calculus (e.g., Fitting, 1996) for clauses. A tableau $T$ for $\mathcal{C}$ is a tree whose non-root nodes are labeled with literals and that fulfills the condition: If the immediate successor nodes $v_1, \ldots, v_n$ of a node $v$ of $T$ are labeled with literals $l_1, \ldots, l_n$, then the clause $\{l_1, \ldots, l_n\}$ (*tableau clause*) is an instance of a clause in $\mathcal{C}$. In the tableau calculus two inference rules are used, namely the *expansion* and the *reduction* rule (e.g., Fitting, 1996). An application of the expansion rule means selecting a clause from $\mathcal{C}$ and attaching the literals of a variant of it to a *subgoal* $s$ which is a literal at the leaf of an *open* branch (a branch that does not contain two complementary literals). Tableau reduction closes a branch by unifying a subgoal $s$ with the complement of a literal $r$ (denoted by $\sim r$) on the same branch, and applying the substitution to the whole tableau.

Connection tableau calculi work on connected tableaux. A tableau is called *connected* or a *connection tableau* if each inner node $v$ (non-leaf node) which is labeled with literal $l$ has a leaf node $v'$ among its immediate successor nodes that is labeled with a literal $l'$ complementary to $l$. The inference rules are *start*, *extension*, and *reduction*. The start rule is always the first inference step of a derivation. It permits a tableau expansion that can only be applied to a trivial tableau, i.e. one consisting of only one node. Note that the start rule can be restricted to so-called *start relevant* clauses without causing incompleteness. Start relevancy of a clause is defined as follows. If $\mathcal{C}$ is an unsatisfiable set of clauses, we call $S \in \mathcal{C}$ start relevant if there is a satisfiable subset $\mathcal{C}' \subset \mathcal{C}$ such that $\mathcal{C}' \cup \{S\}$ is unsatisfiable. Since the set of negative clauses contains at least one start relevant clause, we also consider a restricted calculus which only employs negative clauses for the start expansion ($CTC_{neg}$). The reduction rule is the same as in the conventional tableau calculus. Extension is a combination of expansion and reduction. It is performed by selecting a subgoal $s$ in the tableau $T$, applying an expansion step to $s$, and immediately performing a reduction step with $s$ and one of its newly created successors. Note that in the area of Horn clauses it is sufficient to employ start and extension, i.e. the reduction inference is unnecessary (e.g., Antoniou & Langetepe, 1994). Thus, we assume that we use versions of $CTC$ or $CTC_{neg}$ that do not employ reduction steps in the area of Horn clauses.

$CTC$ or $CTC_{neg}$ do not have specific inference rules for handling equality. Instead, when dealing with equality, the axiomatization must be extended by the reflexivity, symmetry, transitivity, and substitution axioms of the equality symbol. Indeed, the use of an axiomatic form of equality is in no sense optimal. But it is very difficult to develop methods for using built-in equality in tableau calculi that yield convincing results in practice.

If a subgoal $s$ becomes (after some inferences) head of a closed (sub-)tableau we call the obtained substitution a *solution* of $s$.

The notion of a tableau derivation and a search tree is important: We say $T \vdash T'$ if (and only if) tableau $T'$ can be derived from $T$ by applying a start rule if $T$ is the trivial tableau, or by an extension/reduction rule to a subgoal in $T$. The connection tableau calculus is not (proof) confluent. In order to show the unsatisfiability of a clause set $\mathcal{C}$, a search tree, given as follows, has to be examined in a *fair* way (each node of the tree must be visited after a finite amount of time) until a closed tableau occurs. A *search tree* $\mathcal{T}$ defined by a calculus and a set of clauses $\mathcal{C}$ is given by a tree whose root is labeled with the trivial tableau. Every inner node in $\mathcal{T}$ labeled with tableau $T$ has as immediate successors the nodes from the maximal set $\{v_1, \ldots, v_n\}$, where $v_i$ is labeled with $T_i$ and $T \vdash T_i$, $1 \leq i \leq n$.

Since not only the number of proof objects but also their size increases during the proof process, explicit tableaux enumeration procedures that construct all tableaux in $\mathcal{T}$ in a breadth-first manner are not reasonable. Hence, implicit enumeration procedures that apply *consecutively bounded iterative deepening search with backtracking* (Korf, 1985) are normally used. In this approach iteratively larger finite initial parts of the search tree $\mathcal{T}$ are explored with depth-first search. A finite segment is normally defined by a so-called *completeness bound* (which poses structural restrictions on the tableaux which are allowed in the current segment, see below) and a fixed natural number, a so-called *resource*. Iterative deepening is performed by starting with a basic resource value $n \in \mathbb{N}$ and iteratively increasing $n$ until a proof is found within the finite initial segment of $\mathcal{T}$ defined by one bound and $n$. Prominent examples for completeness bounds are the *depth bound*, *inference bound*, and *weighted-depth bound*.

The depth bound limits the maximal depth of inner nodes (non-leaf nodes) in a tableau where the current resource $n$ is the maximal depth permitted (the root node has depth 0). In practice, the depth bound is quite successful (Letz et al., 1994; Harrison, 1996) but it suffers from the large increase of the segment (defined by a resource $n$) caused by an increase of $n$. The inference bound allows a level by level exploration of the search tree (e.g., Stickel, 1988). In comparison with the depth bound, the inference bound makes a smooth increase of the search space possible, but it is inferior to the depth bound in practice. In order to combine the advantages of the depth and the inference bound, the weighted-depth bound was introduced by Moser et al. (1997). This bound describes a class of possible bounds that restrict the tableau depth and the number of inferences allowed to infer a specific tableau.

Goal-orientation of $CTC$, as our typical top-down calculus, is given by the connectedness condition. This condition entails that every literal in a tableau bears a relation to the start clause. The set of possible start clauses can normally be restricted to quite a small set of clauses which are sufficient in order to guarantee completeness (e.g., Moser et al., 1997), e.g. the set of negative clauses as already mentioned. Thus, only certain descendants having a connection to a start (goal) clause are enumerated. A main problem of proofs with $CTC$ is that in general they are rather long. In fact, $CTC$ is among the weakest calculi when the lengths of existing proofs are considered. Therefore, Letz et al. (1994) proposed extensions of $CTC$ which are based on a controlled integration of the *cut rule*. These extensions can also be seen as restricted lemma mechanisms. A further problem is that often during the search tableaux with the same or subsumed subgoals occur repeatedly. There are some extensions of the calculus proposed to overcome such problems. E.g., by Letz et al. (1994) a restricted subsumption concept and by Astrachan and Stickel (1992) caching techniques have been introduced.

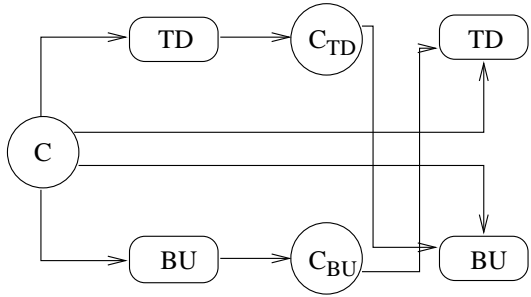

Figure 1: Cooperation between a top-down and a bottom-up prover

## 2.2 Achieving Cooperation by Preprocessing and Input Augmentation

Our approach of integrating top-down and bottom-up provers by cooperation is characterized by *preprocessing* and *input augmentation*. Henceforth, let $\mathcal{C}$ be the initial clause set whose inconsistency should be shown. In the preprocessing phase the bottom-up superposition prover generates a set of clauses $\mathcal{C}_{BU}$ such that $\mathcal{C} \models \mathcal{C}_{BU}$. Analogously, we extract from a certain number of proof attempts of the ME prover a clause set $\mathcal{C}_{TD}$ such that $\mathcal{C} \models \mathcal{C}_{TD}$. Then, the input set $\mathcal{C}$ of the superposition prover is augmented by $\mathcal{C}_{TD}$, and the input of the ME prover with $\mathcal{C}_{BU}$. After that both provers can proceed to work in parallel. Figure 1 displays this kind of cooperation that is essentially based on a sequential concatenation of both provers. The approach can be seen as a specific instantiation of the general cooperation approach TECHS (Denzinger & Fuchs, 1998).

Since the superposition prover works on a search state which contains a set of clauses, it is very easy to generate a set of valid clauses in a preprocessing phase. A very simple method is to perform a fixed number $i$ of activation steps and to generate the clause sets $\mathcal{F}^{A,i}$ and $\mathcal{F}^{P,i}$ of active and passive clauses. Then, we select all *facts*, i.e. positive unit clauses, from $\mathcal{F}^{A,i}$. Since the inferences of a superposition-based prover are sound, it produces only logic consequences of $\mathcal{C}$. As we shall explain in Section 4 in more detail, it is not sensible to add all generated positive units to the input of the top-down prover, i.e. to set $\mathcal{C}_{BU} = \{C : C \text{ is a fact}, C \in \mathcal{F}^{A,i}\}$. However, it is wise to select only some units with a function $\varphi_{BU}$, i.e. $\mathcal{C}_{BU} = \varphi_{BU}(\{C : C \text{ is a fact}, C \in \mathcal{F}^{A,i}\})$.

Because connection tableau-based provers have a search state which contains deductions (tableaux) instead of clauses, it is at first sight not obvious how to extract valid clauses from such a search state which will be well-suited for a superposition-based prover. A common method in order to extract valid clauses is to employ lemma mechanisms of ME provers. Assume that a literal $s$ is a label of the root node of a closed subtableau $T^s$. Let $l_1, \ldots, l_n$ be the literals that are used in reduction steps for closing $T^s$ and that are outside of $T^s$. Then, the clause $\{\sim s, \sim l_1, \ldots, \sim l_n\}$ may be derived as a new lemma (since it is a logical consequence of the tableau clauses in $T^s$). Such a lemma could be transferred to a bottom-up prover. As appealing as this idea sounds, it has some severe restrictions: Such lemmas usually are, e.g. due to instantiations which were previously needed to close other branches, not as general as they could be. Hence, often they cannot be used in inferences, and especially not in contracting inferences which are very important for bottom-up provers. Since these clauses are generated during the proof run in a rather unsystematic way they do not really introduce much goal-orientation and hence do not make use of the advantages of the search scheme typical for ME.

The concept of subgoal clauses permits the generation of clauses derived by inferences involving a proof goal:

**Definition 2.1 (subgoal clause, subgoal clause set)**

1. Let $\mathcal{C}$ be a set of clauses, let $T$ be a tableau for $\mathcal{C}$. A *subgoal clause* $S_T$ regarding $T$ is the clause $S_T = \{l_1, \ldots, l_n\}$, where the literals $l_i$ are the subgoals of the tableau $T$.

2. Let $B$ be a bound, $n \in \mathbb{N}$ be a resource, and $\mathcal{C}$ be a set of clauses.
   If $CTC$ is used, the *subgoal clause set* $\mathcal{S}^{B,n,\mathcal{C}}$ w.r.t. $B$, $n$, and $\mathcal{C}$, is defined by $\mathcal{S}^{B,n,\mathcal{C}} = \{S_T : T$ is a tableau in the initial segment of the search tree for $\mathcal{C}$ and $CTC$ that is defined by $B$ and $n\} \setminus \mathcal{C}$.
   If $CTC_{neg}$ is in use, the *subgoal clause set* $\mathcal{S}^{B,n,\mathcal{C}}_{neg}$ is the set $\mathcal{S}^{B,n,\mathcal{C}}_{neg} = \{S_T : S_T \in \mathcal{S}^{B,n,\mathcal{C}},$ the start expansion of $T$ is performed with a negative clause$\}$.

Note that subgoal clauses are valid clauses, i.e. logical consequences of the initial clause set. The following example illustrates our notion of subgoal clauses.

**Example 2.1** Let $\mathcal{C} = \{\{\neg g\}, \{\neg p_1, \ldots, \neg p_n, g\}, \{\neg q_1, \ldots, \neg q_m, g\}\}$. Then, $\{\neg p_1, \ldots, \neg p_n\}$ is the subgoal clause $S_T$ belonging to the tableau obtained when extending the goal $\neg g$ with the clause $\{\neg p_1, \ldots, \neg p_n, g\}$. If we employ $\mathcal{B} =$ inference bound ($Inf$) and resource $k = 2$, then $\mathcal{S}^{\mathcal{B},k,\mathcal{C}} = \mathcal{S}^{\mathcal{B},k,\mathcal{C}}_{neg} = \{\{\neg p_1, \ldots, \neg p_n\}, \{\neg q_1, \ldots, \neg q_m\}\}$.

A subgoal clause $S_T$ represents a transformation of an original goal clause (which is the start clause of the tableau $T$) into new subgoals realized by the deduction which led to the tableau $T$. The set $\mathcal{S}^{Inf,k,\mathcal{C}}$ is the set of all possible goal transformations into subgoal clauses within $k$ inferences if we consider *all input clauses* to be goal clauses, the set $\mathcal{S}^{Inf,k,\mathcal{C}}_{neg}$ is the set of all possible goal transformations into subgoal clauses within $k$ inferences if we only consider *the negative clauses* to be goal clauses. More exactly, $\mathcal{S}^{Inf,k,\mathcal{C}}$ is the closure of all (goal) clauses w.r.t. (a fixed number $k$ of) extension and reduction steps, $\mathcal{S}^{Inf,k,\mathcal{C}}_{neg}$ is the closure of all negative (goal) clauses w.r.t. $k$ extension and reduction steps.

In order to couple an ME and a superposition prover, we generate in the preprocessing phase with the inference bound and a fixed resource $k > 1$ either the set $\mathcal{S}^{Inf,k,\mathcal{C}}$ or the set $\mathcal{S}_{neg}^{Inf,k,\mathcal{C}}$, depending on whether $CTC$ or $CTC_{neg}$ is used. It is not sensible to set $\mathcal{C}_{TD} = \mathcal{S}^{Inf,k,\mathcal{C}}$ or $\mathcal{C}_{TD} = \mathcal{S}_{neg}^{Inf,k,\mathcal{C}}$ and to augment the input of the bottom-up prover by this set (see Section 3). Thus, we use again a filter function $\varphi_{TD}$ for selecting some subgoal clauses. That is, $\mathcal{C}_{TD} = \varphi_{TD}(\mathcal{S}^{Inf,k,\mathcal{C}}) \subseteq \mathcal{S}^{Inf,k,\mathcal{C}}$ or $\mathcal{C}_{TD} = \varphi_{TD}(\mathcal{S}_{neg}^{Inf,k,\mathcal{C}}) \subseteq \mathcal{S}_{neg}^{Inf,k,\mathcal{C}}$.

Finally, we want to explain how our method—preprocessing and augmentation of the input of the provers—is indeed well-suited for overcoming the disadvantages of the provers.

We start with the top-down prover. Usually the clauses $\mathcal{C}_{BU}$ generated by a superposition prover are quite general because specialized clauses are eliminated by subsumption or rewriting. It is hence quite probable that they can often be used for closing open branches of tableaux enumerated by a top-down prover *without instantiating the tableaux*. If such a kind of lemma matching (e.g., Iwanuma, 1997) is possible we are able to close branches without introducing new subgoals or reducing the possibility that the remaining subgoals are solvable. If subgoals which often occur in tableaux can be solved, the lemmas are a good means for redundancy control. Moreover, since the search scheme of a superposition prover differs from that of an ME prover it is to be expected that it can generate clauses with few inferences that can close a branch which could only be closed by many inferences when using no lemmas. Then proof lengths are drastically reduced. See Section 4 for a more detailed description of the use of lemmas.

The input of a superposition prover is augmented by subgoal clauses which are the result of a transformation of an original goal clause into subgoals. Hence, the goal-orientation of a superposition prover is increased if it uses such transformed goal clauses in its inferences. Since the ME prover employs a different search scheme it can very quickly conduct certain steps that the superposition prover cannot reconstruct because of its heuristic search. The search for a proof can then be reduced.

## 3. Subgoal Clauses for Top-Down/Bottom-Up Integration

In this section we examine the integration of subgoal clauses into the input set of a superposition prover. At first we assume that all subgoal clauses generated within a certain number of inferences are added to the input set and we give some results regarding proof length and search reduction. After that, we explain the necessity of selecting only some subgoal clauses and introduce two selection methods based on the theoretic discussion.

### 3.1 Reduction of Proof Length and Search through Subgoal Clauses

The cooperation method introduced in the preceding section gives rise to the question of whether a *proof length reduction* is possible, i.e. whether there are shorter superposition proofs of the inconsistency of $\mathcal{C} \cup \mathcal{S}^{Inf,k,\mathcal{C}}$ or $\mathcal{C} \cup \mathcal{S}_{neg}^{Inf,k,\mathcal{C}}$ than of the inconsistency of $\mathcal{C}$. (Note that we measure the length of a proof $P$ by counting the number of inference steps $|P|$ in it.) Since some inference steps are necessary for enumerating subgoal clauses we should try to find out whether these inferences can be saved when using the clauses.

This question is mainly of theoretical interest because in general bottom-up provers do not enumerate minimal proofs. Moreover, bottom-up provers usually perform a lot of unnecessary inferences. It is more important to analyze whether a bottom-up prover can benefit from the use of subgoal clauses in the form of a *proof search reduction*, i.e., a reduction of the number of inferences the prover needs in order to find a proof. It is particularly interesting to identify the cases of maximal proof search reduction.

### 3.1.1 Subgoal clause generation via $CTC$

Firstly, we examine the case where we employ calculus $CTC$, i.e. generate $\mathcal{S}^{Inf,k,\mathcal{C}}$. We assume further that no equality is involved in the problem, i.e. superposition corresponds to (ordered) resolution.

**Theorem 3.1**

1. *Let $\mathcal{C}$ be a set of ground clauses not containing equality, let $\square \notin \mathcal{C}$, and let $k > 1$ be a natural number. Let $P_1$ and $P_2$ be minimal length resolution refutation proofs for $\mathcal{C}$ and $\mathcal{C} \cup \mathcal{S}^{Inf,k,\mathcal{C}}$, respectively. Then, it holds: $|P_1| > |P_2|$.*

2. *For each $k > 1$ there is a set of (non-ground) clauses $\mathcal{C}_k$ not containing equality ($\square \notin \mathcal{C}_k$), such that no minimal length resolution refutation proof for $\mathcal{C}_k \cup \mathcal{S}^{Inf,k,\mathcal{C}_k}$ is shorter than a minimal length resolution refutation proof for $\mathcal{C}_k$.*

*Proof:*

1. Note that no factorization steps are needed in the case of ground clauses (recall that clauses are sets of literals). Then, the claim is trivial since the result of the first resolution step of each minimal proof is an element of $\mathcal{S}^{Inf,k,\mathcal{C}}$.

2. Let $k > 1$. Let $\mathcal{C}_k$ be defined by $\mathcal{C}_k = \{\{\neg p(x_1), \ldots, \neg p(x_k)\}, \{p(y_1), \ldots, p(y_k)\}\}$. Let $>= \emptyset$ be the ordering used for ordered resolution. Then, a minimal resolution refutation proof for $\mathcal{C}_k$ requires $k - 1$ binary factorization steps (resulting in the clause $\{p(y_1)\}$) and $k$ resolution steps. Furthermore, it can easily be recognized that there are in $\mathcal{S}^{Inf,k,\mathcal{C}_k}$ only clauses which contain at least one positive and one negative literal. Thus, none of these clauses can lead to a refutation proof for $\mathcal{C}_k \cup \mathcal{S}^{Inf,k,\mathcal{C}_k}$ in less than $2k - 1$ inferences. $\square$

Hence, a reduction of the proof length is at least possible in the ground case. The (heuristic) proof search of a resolution-based prover may not profit from the proof length reduction obtained. For example it is possible that all clauses of a minimal refutation proof of $\mathcal{C}$ have smaller heuristic weights than the clauses from $\mathcal{S}^{Inf,k,\mathcal{C}}$ and will hence be activated before them. Consider following example:

**Example 3.1** Let $>= \emptyset$ be the ordering used for ordered resolution. Let the clause set $\mathcal{C}$ be given by $\mathcal{C} = \{\{\neg a, \neg b, c\}, \{\neg g, b\}, \{a\}, \{g\}, \{\neg c\}\}$. The heuristic $\mathcal{H}$ corresponds to the FIFO heuristic. Furthermore, for the first $n$ activation steps ($n \in \mathbb{N}$, $n \geq 9$), resolvents of the two most recently activated clauses are preferred by $\mathcal{H}$. Then, following clauses are activated by the prover (in this order): $\{\neg a, \neg b, c\}, \{\neg g, b\}, \{\neg a, \neg g, c\}, \{a\}, \{\neg g, c\}, \{g\}, \{c\}, \{\neg c\}, \square$. Furthermore, if the subgoal clauses of $\mathcal{S}^{Inf,k,\mathcal{C}}$ are inserted behind the original axioms the prover will find the same resolution refutation proof for $\mathcal{C} \cup \mathcal{S}^{Inf,k,\mathcal{C}}$ ($k \geq 0$) and the proof search does not benefit from a possible proof length reduction.

Since the above example (especially the chosen heuristic) is somewhat contrived, it can be expected for many problems that clauses from $\mathcal{S}^{Inf,k,\mathcal{C}}$ will be activated. In this case there is also no guarantee that the proof search is improved because the subgoal clauses can import additional redundancy.

$CTC$ differs—apart from the lack of factorization—mainly in the handling of equality from the superposition calculus. In the case that equality is involved in the problem, a proof length reduction even for ground clauses is not guaranteed.

**Theorem 3.2** *For each resource $k > 1$ there is a set of ground unit equations $\mathcal{C}$ ($\Box \notin \mathcal{C}$) where the minimal superposition refutation proofs for $\mathcal{C} \cup \mathcal{S}^{Inf,k,\mathcal{C}}$ are not shorter than minimal proofs for $\mathcal{C}$.*

*Proof:* Let $>= \emptyset$ be the ordering used for superposition. Consider the set of unit equations $\mathcal{C} = \{\{a = b\}, \{f^{k-1}(a) \neq f^{k-1}(b)\}\}$. We assume that $>= \emptyset$ is used as an ordering for superposition. Then, a minimal superposition refutation proof for $\mathcal{C}$ requires two inferences, a superposition step into $f^{k-1}(a) \neq f^{k-1}(b)$ resulting in the inequation $f^{k-1}(a) \neq f^{k-1}(a)$, and then an equality resolution step. In the set $\mathcal{S}^{Inf,k,\mathcal{C}}$ are either non-unit clauses whose refutation requires at least 2 inferences or the units $\mathcal{U} = \{\{f^i(a) \neq f^i(b)\}, \{f^j(a) = f^j(b)\} : 0 \leq i < k-1, 0 < j \leq k-1\}$. Since the refutation of $\mathcal{C} \cup \mathcal{U}$ also requires 2 inferences a reduction of the proof length is impossible. $\Box$

In summary a reduction of the heuristic search for a proof cannot be guaranteed because the proof length may not be reducible, the subgoal clauses may be ignored by the superposition prover, or the subgoal clauses may import too much additional redundancy. The two latter points also hold in the ground case without equality where at least a proof length reduction is guaranteed.

The second point is no real problem since—as our experiments showed—usually subgoal clauses will be activated and will be involved in the search process of a prover. The risk that subgoal clauses are ignored can be minimized by selecting especially such subgoal clauses which will take part in the search with a high probability, e.g. clauses with a small heuristic weight regarding the heuristic of a superposition prover. In our selection process we need not use such a focus according to the experimental results. The first and third point show some theoretical weaknesses but in connection with appropriate relevancy-based selection techniques we could observe in practice that a restructuring of the search caused by using subgoal clauses allows proofs to be found faster.

### 3.1.2 Subgoal clause generation via $CTC_{neg}$

Secondly, we examine the case where we employ the calculus $CTC_{neg}$, i.e. generate $\mathcal{S}^{Inf,k,\mathcal{C}}_{neg}$. Then, even for ground clauses not containing equality it is possible that minimal proofs cannot be shortened when employing subgoal clauses.

**Theorem 3.3** *There is a set of ground clauses $\mathcal{C}$ where no minimal length resolution refutation proof for $\mathcal{C} \cup \mathcal{S}^{Inf,2,\mathcal{C}}_{neg}$ is shorter than a minimal length resolution refutation proof for $\mathcal{C}$.*

*Proof:* Consider the following set $\mathcal{C}$ of clauses (again we employ $>= \emptyset$):

$$\mathcal{C} = \left\{ \begin{array}{ccc} \{l_4, l_6, l_7\}, & \{l_4, l_6, \neg l_7\}, & \{l_3, \neg l_4\}, \\ \{l_3, \neg l_6\}, & \{\neg l_2, \neg l_4\}, & \{l_4, \neg l_5, \neg l_6\}, \\ \{\neg l_2, l_5\}, & \{l_1, l_2, \neg l_3\}, & \{\neg l_1, l_2, \neg l_3\} \end{array} \right\}$$

Each minimal refutation proof for this set requires 9 resolution steps, e.g.:

$$
\begin{array}{llll}
[1, ax] & \{l_4, l_6, l_7\} & [10, res(1, 2)] & \{l_4, l_6\} \\
[2, ax] & \{l_4, l_6, \neg l_7\} & [11, res(6, 10)] & \{l_4, \neg l_5\} \\
[3, ax] & \{l_3, \neg l_4\} & [12, res(3, 10)] & \{l_3, l_6\} \\
[4, ax] & \{l_3, \neg l_6\} & [13, res(7, 11)] & \{\neg l_2, l_4\} \\
[5, ax] & \{\neg l_2, \neg l_4\} & [14, res(5, 13)] & \{\neg l_2\} \\
[6, ax] & \{l_4, \neg l_5, \neg l_6\} & [15, res(4, 12)] & \{l_3\} \\
[7, ax] & \{\neg l_2, l_5\} & [16, res(8, 9)] & \{l_2, \neg l_3\} \\
[8, ax] & \{l_1, l_2, \neg l_3\} & [17, res(14, 16)] & \{\neg l_3\} \\
[9, ax] & \{\neg l_1, l_2, \neg l_3\} & [18, res(15, 17)] & \square
\end{array}
$$

Now, it holds:

$$\mathcal{S}_{neg}^{Inf,2,\mathcal{C}} = \left\{ \begin{array}{ccc} \{\neg l_2, l_6, l_7\}, & \{\neg l_2, l_6, \neg l_7\}, & \{l_1, \neg l_3, \neg l_4\}, \\ \{\neg l_1, \neg l_3, \neg l_4\}, & \{\neg l_2, \neg l_5, \neg l_6\} \end{array} \right\}$$

When enumerating all proofs for $\mathcal{C} \cup \mathcal{S}_{neg}^{Inf,2,\mathcal{C}}$ one can recognize that the minimal proof length cannot be reduced. $\square$

In the case where equality is involved in our proof problems, we obtain a theorem analogous to the previous one:

**Theorem 3.4** *For each resource $k > 1$ there is a set of unit equations $\mathcal{C}$ where the minimal superposition refutation proofs for $\mathcal{C} \cup \mathcal{S}_{neg}^{Inf,k,\mathcal{C}}$ are not shorter than minimal proofs for $\mathcal{C}$.*

*Proof:* In analogy to Theorem 3.2. $\square$

We can recognize that for $CTC_{neg}$ the results are essentially the same as for $CTC$. In general the reduction of the heuristic search for a proof cannot be guaranteed and proof lengths may not always be reduced. In practice, however, we could again observe that a restructuring of the search caused by subgoal clauses often allows proofs to be found faster.

In the following we simply assume that inferences with subgoal clauses are not omitted by a superposition prover, i.e. that they are involved in the proof search. We want to deal in more detail with the problem of identifying subgoal clauses which can lead to a large reduction of the search effort and how we can efficiently select such subgoal clauses. This problem has to be tackled with heuristics since there is—as we have examined—no theoretical guarantee and also no method to decide whether subgoal clauses are useful.

## 3.2 Relevancy-Based Generation of Subgoal Clauses

Even when using small resources $k$ the sets $\mathcal{S}^{Inf,k,\mathcal{C}}$ and $\mathcal{S}_{neg}^{Inf,k,\mathcal{C}}$ can become quite large. Thus, it is not sensible to integrate all subgoal clauses from $\mathcal{S}^{Inf,k,\mathcal{C}}$ or $\mathcal{S}_{neg}^{Inf,k,\mathcal{C}}$ into the search state of a superposition-based prover. Integrating too many clauses usually does not entail a favorable rearrangement of the search because the heuristic "gets lost" in the huge number of clauses which can be derived from many subgoal clauses. It is hence reasonable to develop techniques for filtering subgoal clauses that appear to entail a large gain in efficiency for a superposition prover if they can be proven. That is, we are interested in filtering *relevant* subgoal clauses. As already described in Section 2, we generate a *set of subgoal clause candidates* and then we select some subgoal clauses from this set. The chosen subgoal clauses are added to the search state of the bottom-up prover. In the following, we will at first introduce some criteria for measuring the relevancy of a clause. Then, we shall introduce two techniques for generating subgoal clause candidates and deal with the selection of relevant subgoal clauses.

### 3.2.1 Relevancy Criteria for Subgoal Clauses

Two main characteristics of subgoal clauses can contribute to a speed-up of the proof search.

Firstly, since according to Section 3.1 subgoal clauses introduce additional redundancy it is important that some of the clauses can be proven quite easily, that is more easily than the original goal(s). In order to estimate this, it is necessary to judge whether they can *probably be solved* with the help of clauses of the input set. Measuring similarity between a goal and other clauses with the techniques developed by Denzinger and Fuchs (1994), Denzinger et al. (1997), or Fuchs (1997) may be well-suited for this estimation.

Secondly, a solution of a newly introduced subgoal clause *should not always entail a solution of an original goal within few steps of the superposition-based prover.* If this were the case then the integration of new subgoal clauses would not promise much gain. Criteria in order to estimate this are: Firstly, the transformation of an original goal clause into a subgoal clause by an ME prover should have been performed by using many inferences, i.e. $k$ should be quite high. Then, a solution of a new subgoal clause usually does not entail a solution of an original goal within few steps because the probability is rather high that a bottom-up prover cannot—due to its heuristic search—quickly reconstruct the inferences needed to infer the original goal. Secondly, if there is a subgoal clause $S_T$ and some of the tableau clauses of the tableau $T$ have a high heuristic weight w.r.t. the heuristic of the superposition-based prover, a high gain of efficiency can be expected if the prover can prove $S_T$. This is due to the fact that inferences needed to infer the original goal using $S_T$ are difficult for the superposition-based prover.

### 3.2.2 Efficient Generation and Selection of Subgoal Clauses

In order to generate a set of interesting subgoal clauses it is important that we employ a large resource for generating subgoal clauses. As we have already discussed, subgoal clauses that are generated with a small number of inferences do not promise much gain because a bottom-up prover may easily reconstruct the inferences needed to infer them. However, it is not possible to generate all subgoal clauses $\mathcal{S}^{Inf,k,\mathcal{C}}$ or $\mathcal{S}^{Inf,k,\mathcal{C}}_{neg}$ for a sufficiently large resource $k$ as subgoal clause candidates because their huge number means that the costs of generation and additional selection are too high. Hence, we must restrict ourselves to a set of subgoal clause candidates that is a *subset* of $\mathcal{S}^{Inf,k,\mathcal{C}}$ or $\mathcal{S}^{Inf,k,\mathcal{C}}_{neg}$, $k$ sufficiently large (see Section 5).

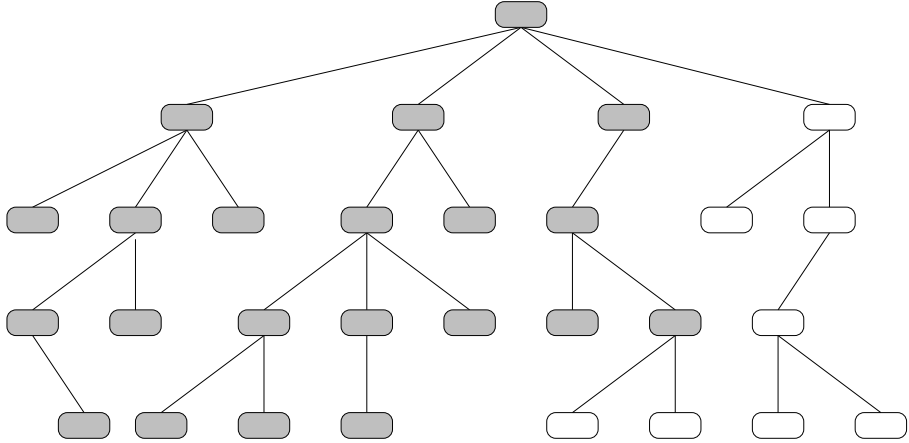

Figure 2: Inference-based generation of a set of subgoal clause candidates

Our first variant, an *inference-based* method, starts by generating subgoal clauses from $\mathcal{S}^{Inf,k,\mathcal{C}}$ or $\mathcal{S}_{neg}^{Inf,k,\mathcal{C}}$ for a rather large resource $k$ and stops when $N_{sg}$ subgoal clause candidates are generated. The advantage of this method is that it is very easy and can be efficiently implemented. Tableaux are enumerated with a fixed strategy for selecting subgoals for inferences (usually left-most/depth-first) and for each tableau its subgoal clause is stored. The main disadvantage of this method is that due to the fixed strategy and the limit of the number of subgoal clauses, we only obtain subgoal clauses which are inferred from goal clauses by expanding particular subgoals with a large number of inferences and other subgoals with only a small number of inferences. (See also Figure 2: Ovals are tableaux in a finite segment of the search tree $\mathcal{T}$, the lines represent the $\vdash$ relation. Grey ovals represent enumerated tableaux, i.e. their subgoal clauses are stored, white ovals represent tableaux which are not enumerated.) Thus, the method is somewhat unintelligent because no information about the quality of the transformation of an original goal clause into a subgoal clause is used. Certain transformations are favored against others only due to the uninformed subgoal selection strategy.

Our second variant, an *adaptive* method, tries to overcome these disadvantages in the following way: Instead of permitting more inferences when generating subgoal clauses due to an uninformed subgoal selection strategy, we want to allow more inferences at certain *interesting positions* of the search tree $\mathcal{T}$ for a given set of clauses $\mathcal{C}$.

In detail, our approach is as follows: At first, we generate all subgoal clauses $\mathcal{S}^{Inf,k_1,\mathcal{C}}$ or $\mathcal{S}_{neg}^{Inf,k_1,\mathcal{C}}$ with a resource $k_1$ which is smaller compared to the first variant. Then, a fixed number $N_{ref}$ of subgoal clauses is chosen which promise the highest gain of efficiency regarding the previously mentioned criteria. More exactly, we choose subgoal clauses which are maximal w.r.t. a selection function $\psi$. One possible realization of $\psi$ is:

$$\psi(S_T) = \alpha_1 \cdot I(S_T) + \alpha_2 \cdot \max(\{\mathcal{H}(C) : C \text{ is a tableau clause in } T\}) \\ + \alpha_3 \cdot \max(\{sim(S_T, C) : C \in \mathcal{C}, |C| = 1\})$$

The higher the number of inferences $I(S_T)$ which are needed to infer $S_T$, the higher $\psi(S_T)$ should be. Hence, $\alpha_1$ should be positive. Setting $\alpha_2 > 0$ is also sensible. If there are tableau clauses in $T$ which have a high heuristic weight regarding the heuristic $\mathcal{H}$ of the superposition-based prover we can—as already discussed—gain a lot of efficiency. The function $sim$ measures whether literals from $S_T$ can probably be solved with unit clauses from $\mathcal{C}$. It maps a pair of clauses to a real number. The larger $sim(S_T, C)$ the larger the similarity between $S_T$ and the unit clause $C$. We utilize a variant of the function $occnest$ which is defined by Denzinger and Fuchs (1994) for accomplishing the task. We set $\alpha_1 > \alpha_2 > \alpha_3 \geq 0$ due to the increasing vagueness of the criteria.

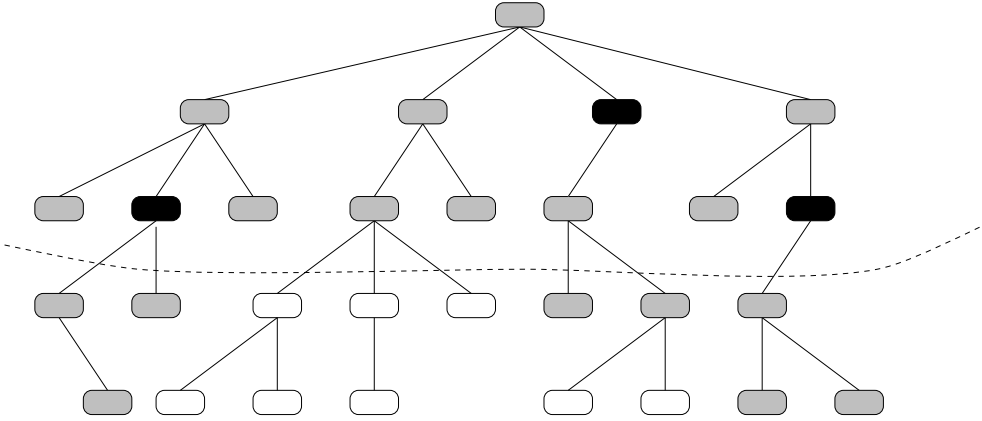

Figure 3: Adaptive generation of a set of subgoal clause candidates

Now, let $M^{N_{ref}} \subseteq \mathcal{S}^{Inf,k_1,\mathcal{C}}$ or $M^{N_{ref}} \subseteq \mathcal{S}^{Inf,k_1,\mathcal{C}}_{neg}$ be the set of chosen subgoal clauses. Then, we generate subgoal clauses with a resource $k_2$ but employ the clauses from $M^{N_{ref}}$ as start clauses for the subgoal clause enumeration. We call the set of subgoal clauses generated with this method $\mathcal{S}^{Inf,k_2,\mathcal{C},M^{N_{ref}}}$. (Consider also Figure 3: The dotted line shows which subgoal clauses are generated with resource $k_1$. Then some of them are selected (black ovals) and used as starting points for the generation of new subgoal clauses with resource $k_2$.) The resource $k_2$ should again not be too high in order to allow a fast enumeration of the subgoal clauses. The set of subgoal clause candidates is then given by $\mathcal{S}^{Inf,k_1,\mathcal{C}} \cup \mathcal{S}^{Inf,k_2,\mathcal{C},M^{N_{ref}}}$ (if we employ $CTC$) or $\mathcal{S}^{Inf,k_1,\mathcal{C}}_{neg} \cup \mathcal{S}^{Inf,k_2,\mathcal{C},M^{N_{ref}}}$ (if we employ $CTC_{neg}$). Thus, subgoal clause candidates are on the one hand all subgoal clauses derived with a certain number $k_1$ of inferences such that it can be assumed that the proof length is reduced. On the other hand, we have some subgoal clause candidates which are derived with a higher number of inferences, at most $k_1 + k_2$. These subgoal clauses promise a high gain of efficiency because they are derived from subgoal clauses selected with function $\psi$. That is, they are derived from clauses which are considered to be very relevant for a superposition-based theorem prover.

For selecting subgoal clauses from the set of subgoal clause candidates we employ function $\varphi$—which is mainly based on the function $\psi$—and select clauses with the highest weight regarding $\varphi$. $\varphi$ is defined by

$$\varphi(S_T) = \psi(S_T) - \theta(S_T).$$

$\theta$ simply counts a weighted sum of the number of variables in $S_T$ and two times the number of function or predicate symbols in $S_T$. Hence, quite "general" subgoal clauses are preferred. This is sensible because they can usually be solved more easily.

## 4. The Use of Lemmas in Model Elimination

In this section we examine theoretical and practical aspects concerning the integration of superposition-generated lemmas into the input set of a model elimination prover. At first, we present some results regarding proof length and search reductions. As before, we measure the proof length by the number of inferences in it. Proof search is measured by the number of inferences the prover must perform in order to find a proof. Then, we introduce some methods for a relevancy-based filtering of lemmas.

### 4.1 Proof Length Reduction

When adding positive unit lemmas $\mathcal{C}_{BU}$ of a bottom-up prover to the axiomatization $\mathcal{C}$ of a top-down prover, a proof length reduction is possible if the following situation occurs. If $T$ is a tableau that represents a proof and $sg$ is a literal which has a depth smaller than $n-1$ of a branch in $T$ with depth $n$ we can reduce the proof length if $\sim sg$ is unifiable with a lemma $l \in \mathcal{C}_{BU}$. Hence, we are interested in the question of whether we can find lemmas useful in the described sense if we choose $\mathcal{C}_{BU} = \varphi_{BU}(\{C : C \text{ is a fact}, C \in \mathcal{F}^{A,i}\})$ as proposed in Section 2. Unfortunately, even if $\varphi_{BU}$ selects all facts in $\{C : C \text{ is a fact}, C \in \mathcal{F}^{A,i}\}$ and $i$ is arbitrary large there is no guarantee that we can find a useful lemma in this set. This is even true if the bottom-up prover employs a fair heuristic.

**Theorem 4.1** *For each $i \in I\!N$ there is a clause set $\mathcal{C}_i$ and a fair heuristic $\mathcal{H}_i$ such that no positive unit lemma from $\mathcal{F}^{A,i}$ (generated by a resolution prover starting with $\mathcal{C}_i$ and employing heuristic $\mathcal{H}_i$) can reduce the proof length of a proof for $\mathcal{C}_i$ with CTC ($CTC_{neg}$).*

*Proof:* Let $i$ be a natural number and $>= \emptyset$ be the ordering used for ordered resolution. For a literal $l$, $|l|$ denotes the number of symbols in $l$. Set $\mathcal{C}_i = \{\{p(a)\}, \{\neg p(x), p(f(x))\}, \{\neg q(a)\}, \{\neg q(b), q(a)\}, \{q(b)\}\}$. Let $\mathcal{H}_i(\{l_1, \ldots, l_n\}) = \sum_{j=1}^{n} \mathcal{H}_i^L(\bar{l}_j)$, with $\bar{l}_j = l_j$, if $l_j$ is positive, and $\bar{l}_j = \sim l_j$, otherwise. Moreover,

$$\mathcal{H}_i^L(l) = \begin{cases} |l| & , l \equiv p(t), t \text{ is a term} \\ 2 + i + |l| & , l \equiv q(t), t \text{ is a term.} \end{cases}$$

Then it holds: For a fixed parameter $i$, $\mathcal{H}_i$ is a fair heuristic. Moreover, there are only ME proofs of the inconsistency of $\mathcal{C}_i$ which contain literals with top-symbol $q$. $\mathcal{F}^{A,i}$ contains only literals with top-symbol $p$, though. Hence, it is impossible that a lemma of $\mathcal{F}^{A,i}$ is applicable. $\square$

From a theoretical point of view we have again the negative result that in general useful lemmas are not elements of $\mathcal{F}^{A,i}$. However, empirical studies (see Section 5) reveal that in the most cases useful lemmas are generated by a superposition prover. Hence, we assume that useful lemmas are in $\mathcal{C}_{BU}$ and henceforth examine which lemmas—being part of a proof—can lead to a high proof search reduction of an ME prover.

## 4.2 Proof Search Reduction

The effects regarding the structure of the search space of a $CTC$ ($CTC_{neg}$) prover caused by the use of lemmas are closely related to the *utility problem* (e.g., Minton, 1990) from the area of explanation-based learning (EBL) and macro operator learning (see also Markovitch & Scott, 1993). At the first sight, lemma use could be interpreted as introducing new edges into the original search tree $\mathcal{T}$ because a sub-deduction (proof of a lemma) can be reduced to one inference by applying a lemma. This corresponds to macro operator learning or EBL where inference chains are generalized and disjunctively stored as new operators or concept descriptions (e.g., Minton, 1990), respectively. We should notice, however, that the use of lemmas does not only insert new edges but also new *nodes* into the search tree. This comes from the fact that the structure of a tableau $T_1$ where a lemma is applied differs from the structure of an in other parts equal tableau $T_2$ where the lemma proof is "expanded". This has no influence on the inferences possible with $T_1$ and $T_2$ (the edges outgoing from the nodes $v_1$ and $v_2$ that are labeled with $T_1$ and $T_2$, respectively) but it has an effect on the value a completeness bound $\mathcal{B}$ assigns to the tableaux. Considering the bounds introduced in Section 2, $T_1$ can be enumerated with a resource value which is smaller than or equal to that needed to enumerate $T_2$. In analogy to macro operator learning and considering these remarks, we now summarize the advantages and disadvantages of using lemmas *in connection with iterative deepening procedures*.

A minor advantage of introducing a lemma is the advantage of *decreasing path costs*, i.e. the costs of reproducing the inferences needed for its proof. The major advantage of using lemmas is that they make a *restructuring of the the search space possible*.

On the one hand, one can save the possibly high search effort needed for proving a useful lemma (assuming the lemma proof can be expanded within the finite segment of $\mathcal{T}$ to be considered). On the other hand, it is possible that a closed tableau can be reached within a smaller resource value ("resource reduction"). Then, the reordering effects usually allow us to solve problems that were previously out of reach because the search procedure gets lost in the (usually exponentially) larger segment of the search tree defined by a larger resource. It is clear, however, that this advantage only holds if the segment of the search tree defined by the lower resource value is not increased too much by the use of the lemmas. Considering our search bounds we can see that normally resource reductions cannot be guaranteed when using superposition generated lemmas in an ME prover. When using the inference bound a resource reduction is guaranteed if by using lemmas a proof length reduction can be obtained. In the case of the depth and weighted-depth bound in general not even a proof length reduction leads to a resource reduction.

Besides the positive effects of using lemmas, some negative effects also occur. These stem from an increased redundancy. The main disadvantage regarding the use of lemmas is the increase of the branching rate of the search tree. It is possible that a *misleading solution* of a subgoal may be obtained that could not be found before within a given finite segment. Even if a resource reduction from $n$ to $n' < n$ occurs it is possible that solutions of subgoals that could not be found with resource $n$ (without lemmas) can now be found with resource $n'$ and lemmas. This can reorder the search space in a hardly controllable manner. Considering the inference bound in some cases tableaux which could not be enumerated with resource $n$ can now be enumerated with lemmas. It is possible that the use of lemmas

"spares" more than $n - n'$ inferences. If we use the depth bound it might be that in a tableau some branches which can be closed in a depth greater than $n$ without lemmas can now be closed in a depth smaller than or equal to $n'$. Then a lot of superfluous inferences can be introduced to the new minimal proof segment. Analogous effects take place when using the weighted-depth bound (especially also when using the configuration of the bound as described by Moser et al. (1997)).

Additionally, *duplications of segments of the search space* can occur. Assuming the expanded lemma proof lies within the initial segment of $\mathcal{T}$ to be considered, the use of an irrelevant lemma can cause a repeated exploration of parts of the search space which does not contain a proof: Since a superfluous solution of a subgoal is found twice (via the lemma and by performing the inferences needed to prove the lemma) the resulting superfluous inferences have to be performed twice, too. This disadvantage, however, can usually be overcome by using local failure caching (Letz et al., 1994).

Besides these effects, which cause a restructuring of the search space, it is even possible that the use of lemmas *increases the number of solutions* of certain subgoals that exists in the whole search tree. This is because the use of lemmas can shadow well-known pruning techniques like regularity since no regularity checks are possible in the proof of a lemma. Furthermore, the newly introduced lemmas cause the problem that in each inference a possibly large number of lemmas has to be tested in order to determine whether inferences are possible (*applicability test*). This necessitates new unification attempts.

In summation our lemma mechanism is in general not able to produce lemmas that lead to a proof length reduction and thus to a resource reduction. Nevertheless experience shows that in many cases reductions of the proof length and the needed resource can be obtained. When a small number of lemmas is sufficient for a resource reduction the number of inferences which can be spared by using lemmas can exceed the number of new superfluous inferences by magnitudes. Thus, mechanisms are needed in order to select *relevant* lemmas from $\{C : C \text{ is a fact}, C \in \mathcal{F}^{A,i}\}$ that should be inserted into $\mathcal{C}_{BU}$. If we can find a rather small lemma set which permits a resource reduction then in almost all cases we can find a proof much faster than would be possible without using lemmas (see Section 5). Note that the case of a large reduction of the proof length without a reduction of the resource needed normally performs significantly worse than the case of a small proof length reduction with a reduction of the resource needed.

## 4.3 Relevancy-Based Lemma Selection

Analogous to the foregoing section we now want to introduce some abstract principles for filtering lemmas based on the discussion regarding the structure of the search space. Then, we deal with concrete heuristics applied for selecting lemmas.

### 4.3.1 RELEVANCY CRITERIA FOR LEMMAS

Since superposition provers employ a different search scheme than ME provers and since they have effective mechanisms for handling equality, we can assume that a few subgoals which are hard to solve (the proof necessitates a large resource w.r.t. a given completeness bound) with an ME prover can be solved with lemmas. However, when using lemmas in order to close some subgoals of an open tableau the remaining open subgoals should be easy to solve w.r.t. the given bound. Otherwise, we still cannot solve the problem within a smaller resource. Note that we usually cannot expect that all branches of a proof can be "shortened" by superposition-generated lemmas. Since our lemma generation provides no guarantee that useful lemmas are generated (see Section 4.1) usually only a small number of lemmas can be employed in a proof. All in all we obtain that "interesting" proofs (i.e. those we want to find) for an application of lemmas are proofs that contain many subgoals that are easy to solve—and can hence be solved "conventionally" within a small resource—and only a few hard subgoals that must be solved with lemmas. Then, we can expect that using lemmas leads to resource reductions. Our filter techniques should hence try to find lemmas that might be part of such proofs.

Furthermore, we should estimate how many new superfluous inferences are introduced by a lemma. The integration of new lemmas must not increase the branching rate too much. Otherwise, the gain of a possible resource reduction is negated by the large overhead (see Section 4.2).

These criteria lead us to three different filter functions that concentrate on certain aspects of relevancy. The filter functions are well-suited for all of the previously introduced completeness bounds. Due to the vagueness of the filter criteria we use each filter function in order to choose some lemmas (see Section 5). Note that it is better to select a few unnecessary lemmas than to omit the selection of important ones.

### 4.3.2 EXPANSION/CONTRACTION-BASED SELECTION

The first filter function is called $\varphi_{BU}^S$. This function is rather simple and aims at finding lemmas that do not lead to a high increase of the branching rate. $\varphi_{BU}^S$ accomplishes this by using knowledge obtained in the lemma generation (preprocessing) phase of the bottom-up prover. In detail, $\varphi_{BU}^S$ selects facts with the highest value regarding a judgment function $\psi_{BU}^S$.

**Definition 4.1 (Judgment function $\psi_{BU}^S$)**
For a positive unit $l$, generated in the preprocessing phase, let $\varepsilon(l)$ and $\kappa(l)$ be the numbers of expansion and contraction inferences, respectively, that $l$ was involved in. Then

$$\psi_{BU}^S(l) = \kappa(l) - \varepsilon(l).$$

$\psi_{BU}^S$ counts the inferences that each lemma candidate was involved in and rates expanding inferences negative, contracting inferences positive. If a fact $l$ was often involved in an expanding inference like resolution or superposition, then $l$ or many subterms of it are unifiable with (subterms of) (maximal) literals of axioms or derived clauses. Hence, if $l$ is added to the axiomatization of the top-down prover it can be expected that $l$ or certain descendants of it can very often take part in extension steps. Since this leads to a high increase of the branching rate we rate this negative. Of course, we are interested in the fact that a lemma can be applied by the ME prover. However, since lemmas of a superposition prover are usually quite general most of them may be used for closing occurring subgoals.

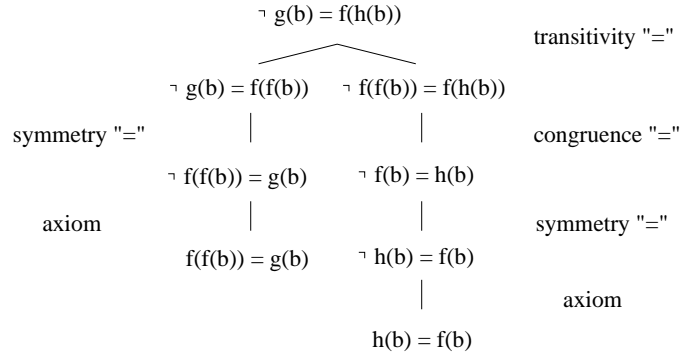

Figure 4: Simulating a superposition step with ME

Our criterion aims mainly at excluding lemmas that are applicable in too many cases and hence introduce too many solutions of subgoals that do not lead to a refutation of the input clauses. In contrast to expanding inferences we rate contraction inferences positively. Indeed, ME provers do not have contracting inferences. But as shown by Letz et al. (1992), subsumption can partly be simulated by *subsumption constraints*. Hence, clauses that are able to contract many other clauses can support search pruning techniques.

### 4.3.3 Derivation-Based Selection

The second filter function $\varphi_{BU}^D$ tries to select facts that are able to close subgoals that are very hard to solve with a connection-tableau-based prover. In order to estimate this, we consider the derivation history of a fact. We employ this filter function only if equality is involved in a problem.

**Example 4.1** Let $\{f(f(x)) = g(x)\}$ and $\{h(b) = f(b)\}$ be axioms. Then, the clause $\{g(b) = f(h(b))\}$ can be derived by one superposition step. However, if $\neg g(b) = f(h(b))$ is a subgoal of an ME proof, its proof is more complicated (see Figure 4).

In general, if the superposition step is performed at a position $p$ and $|p|$ denotes the depth of the position (above $|p| = 1$), then at most $|p| + 5$ inferences are needed in order to prove the result of such a superposition step. The proof necessitates at most a depth of $|p| + 3$.

This example shows that the simulation of the specific equational operations of a superposition prover necessitates a high depth as well as inference resource in an ME prover. That is, lemmas derived by many superposition steps are able to close subgoals that cannot be solved by an ME prover within small resources. Hence, if such lemmas are applicable large resource reductions possibly occur for depth or inference oriented bounds. The judgment function $\psi_{BU}^D$ employs this criterion. Again, the filter function $\varphi_{BU}^D$ selects facts with the highest value regarding this judgment function.

**Definition 4.2 (Judgment function $\psi_{BU}^D$)**

For a positive unit $l$ generated in the preprocessing phase, let $\psi_{BU}^D(l)$ be defined by

$$\psi_{BU}^D(l) = \begin{cases} 0 & , l \text{ is an axiom} \\ \psi_{BU}^D(l_1) + \psi_{BU}^D(l_2) + 1 & , l \text{ is derived by a superposition step with premises} \\ & \quad l_1 \text{ and } l_2 \\ \sum_{i=1}^n \psi_{BU}^D(l_i) & , l \text{ is derived by a non-superposition inference} \\ & \quad \text{involving the literals } l_1, \ldots, l_n \ (n \in \mathbb{N}). \end{cases}$$

### 4.3.4 Complexity-Based Selection

Our third filter function $\varphi_{BU}^C$ aims at selecting lemmas that are able to solve some hard subgoals of ME subgoal clauses such that the resulting open subgoals can easily be solved. Hence, the judgment function $\psi_{BU}^C$ used by $\varphi_{BU}^C$ considers the sets of subgoal clauses $\mathcal{S}^{\mathcal{B},n,\mathcal{C}}$ or $\mathcal{S}_{neg}^{\mathcal{B},n,\mathcal{C}}$ for a certain resource $n$. For each subgoal clause $sg$, if a lemma $l$ with $\psi_{BU}^{C,sg}(l) > 0$ exists, the lemma $l$ with the highest judgment $\psi_{BU}^{C,sg}(l)$ is selected until a maximal number of lemmas is selected (see Section 5). This judgment is computed as follows.

**Definition 4.3 (Judgment function $\psi_{BU}^{C,sg}$)**

For a positive unit $l$ generated in the preprocessing phase and a subgoal clause $sg = \{l_1, \ldots, l_m\}$, let $\psi_{BU}^{C,sg}(l)$ be defined in the following manner:

If no subgoal can be solved with $l$, i.e. $\neg \exists i, 1 \le i \le m, \sigma : \sigma = mgu(\sim l_i, l)$, then $\psi_{BU}^{C,sg}(l) = 0$. Otherwise, let $sg_U = \{u_1, \ldots, u_k\} \subseteq sg$ be a set of literals and $\sigma$ be a substitution so that $\sigma$ is most general with: $\forall z, 1 \le z \le k : \sigma(\sim u_z) \equiv \sigma(l)$. Moreover, under all subsets of $sg$ and substitutions with this property, let the set $sg_U$ and the substitution $\sigma$ be a maximum of the function $G^{sg}$, defined by $G^{sg}(\{u_1, \ldots, u_k\}, \sigma) = \frac{k}{1 + \sum_{z=1}^k |\sigma(u_z)| - |u_z|}$. [1] Then, the remaining literals of $sg$ are $sg_R = \{r_1, \ldots, r_j\} = sg \setminus sg_U$. Let $\gamma$ be a complexity function, i.e. $\gamma$ maps literals to $[0; 1]$ and high values of $\gamma$ indicate that the respective literal (subgoal) appears to be solvable. Then,

$$\psi_{BU}^{C,sg}(l) = \sum_{r \in sg_R} \gamma(\sigma(r)) - \sum_{u \in sg_U} \gamma(\sigma(u)) - j.$$

We can recognize that $\psi_{BU}^{C,sg}(l)$ really rates $l$ with a high value if many hard subgoals (w.r.t. $\gamma$) of $sg$ can be solved with $l$. Moreover, $l$ is rated with a high value if there are only a few subgoals in the subgoal clause $sg$ that cannot be solved by lemmas and that appear to be solvable rather easily (w.r.t. $\gamma$). In our realization, $\gamma$ considers subgoals to be solvable that are small, have a rather flat term structure, and many variables in comparison with the term size. In future we will further refine $\psi_{BU}^{C,sg}$ by explicitly considering the completeness bound which is used for the top-down proof search.

---

1. $sg_U$ is a rather large set of subgoals that can be solved with $l$ such that not too many symbols are introduced by the unifier. This is sensible because otherwise the possibility that the remaining subgoals can be solved is decreased too much.

## 5. Experimental Study

In order to conduct an experimental evaluation of our integration of top-down/bottom-up provers, we coupled two renowned provers: the ME prover SETHEO and the superposition prover SPASS. We have used the version of SETHEO as described by Moser et al. (1997). SPASS has been employed in version 0.55.

### 5.1 Architecture and Behavior of the Experimental System

Our experimental environment can be described as follows: Each prover runs on its own processor and obtains the initial clause set $\mathcal{C}$ as input. We employed a rather efficient method to organize the preprocessing. Essentially, the top-down prover generates subgoal clauses with one of the two variants. In our environment this does not require changes in the top-down prover but can be performed with built-ins of the PROLOG-style input language of SETHEO. Since SETHEO employs $CTC_{neg}$ we experimented only with subgoal clauses obtained with negative start clauses. Then, these subgoal clauses are filtered, transferred to the bottom-up prover, and integrated into its search state. The preprocessing of the bottom-up prover is performed in parallel with the preprocessing of the top-down prover. The prover activates clauses with its basic heuristic until the top-down prover finishes its preprocessing. Thus, we achieve synchronization of the provers. After that, we extract the positive units from the set of active facts of SPASS and filter some facts as described. Since we can employ the generated subgoal clauses of SETHEO for the filter function $\varphi_{BU}^{C}$ the generated subgoal clauses can be used as additional input of SPASS as well as for the selection of lemmas for SETHEO. Finally, the provers proceed to tackle the problem in parallel with their standard settings. By using this environment we can achieve cooperation by exchanging lemmas and subgoal clauses without one concept disturbing the other. In contrast, both concepts support each other because results obtained from one preprocessing can be employed in the other.

We experimented in the light of problems stemming from the well-known problem library TPTP v.1.2.1 (Sutcliffe et al., 1994; Sutcliffe & Suttner, 1998). In order to obtain a reliable collection of data, we employed all domains contained in the TPTP as our test set. Because these domains cover a wide range of very different problems we assume that this is a reliable test set.

Since the TPTP contains too many problems to list and discuss the runtimes of single problems, we will present an overview of the number of solved problems in the TPTP library. Furthermore, we study in which domains cooperation is especially important and deal with the main features responsible for this. In addition we study the results in three domains in more detail to give an impression for the decrease in run time. This concerns the domains CAT (category theory), LDA (LD-algebras), and COL (combinatory logic). The problems in the domains CAT and LDA contain equality as well as non-Horn clauses. COL is a Horn-equality domain.

In detail, the parameters of our experimental system are: The subgoal clause candidates were generated in such a way that for variant 1 we employed the resource $k = 10$ which performed best in the experiments. The use of higher resources did not yield better results. We limited the set of subgoal clauses by $N_{sg} = 500$. For variant 2 we employed $k_1 = k_2 = 9$ as resources. As start clauses for an adaptive refinement we selected $N_{ref} = 5$ subgoal clauses. These parameters allowed the efficient generation of all subgoal clauses within the initial segments of the search tree. Usually with this method at most 500 subgoal clauses were generated, i.e. about the same number as when employing variant 1. For the selection of subgoal clauses that are to be transmitted to Spass we used domain-dependent parameters. For CAT, COL, and LDA we used 100 clauses. In the other domains in preliminary experiments the use of 30 clauses achieved the best results.

The bottom-up lemmas were selected via the functions $\varphi_{BU}^S$, $\varphi_{BU}^D$, and $\varphi_{BU}^C$. We selected with each of the functions a maximum of 10 clauses.

The setting of Setheo was automatically chosen as described by Moser et al. (1997). The Spass standard heuristic essentially selects clauses of the smallest size. Periodically, clauses are selected with breadth-first search.

## 5.2 Experimental Results

In the following we compare the results of our cooperative prover with the single provers. This comparison is performed regarding the whole TPTP library. After that, we analyze runtimes in few selected domains of TPTP.

### 5.2.1 Comparison of different variants in the TPTP

Table 1 presents results of our experiments. It shows the number of solved hard problems in certain domains of TPTP. Solved means that a proof could be found within 300 seconds. We consider a problem to be hard if neither Spass nor Setheo are able to solve it within 10 seconds. The table only presents the results of such domains where hard problems exist and where at least one hard problem could be solved by any of the considered variants. Note that the table cannot give hints on the power of the single provers. This is because it does not give the complete number of problems which can be solved by each single prover in the whole domain. Since many non-hard problems are in the TPTP this number is usually much higher than the number of solved hard problems. Nevertheless, the table is sufficient for analyzing the performance of our cooperative system since only the hard problems are interesting for studying the potential of cooperation.

Column 1 of the table displays the name of the domain. Columns 2 and 3 present the number of solved problems of Spass and Setheo (on a SPARCstation-20/712) when working alone. Column 4 shows the number of solved problems of Spass when it obtains subgoal clauses from Setheo which are generated regarding variant 2. This variant performs better than variant 1 (see also the following subsection). Column 5 displays the number of solved problems of Setheo when it obtains bottom-up generated lemmas from Spass. In that case we always employed variant 2 for generating subgoal clauses (recall that the selection of lemmas depends on the way how subgoal clauses are generated). Column 6 gives the number of solved problems of a competitive version of Spass and Setheo in order to show that our cooperative prover is indeed much more powerful than a simple competitive parallel prover. Finally, in column 7 we can find the number of solved problems of our cooperative system.

| domain | Spass | Setheo | adaptive | lemma | competitive | cooperative |
|--------|-------|--------|----------|-------|-------------|-------------|
| ANA | 0 | 2 | 2 | 2 | 2 | 4 |
| BOO | 4 | 3 | 5 | 4 | 5 | 6 |
| CAT | 6 | 4 | 11 | 8 | 7 | 14 |
| CIV | 1 | 0 | 1 | 0 | 1 | 1 |
| COL | 2 | 10 | 2 | 14 | 12 | 16 |
| GEO | 7 | 10 | 8 | 11 | 14 | 15 |
| GRP | 30 | 1 | 36 | 3 | 30 | 38 |
| HEN | 7 | 5 | 9 | 7 | 10 | 12 |
| LCL | 25 | 8 | 26 | 9 | 30 | 31 |
| LDA | 5 | 1 | 9 | 1 | 6 | 9 |
| NUM | 1 | 0 | 1 | 0 | 1 | 1 |
| PLA | 3 | 2 | 3 | 2 | 3 | 3 |
| RNG | 4 | 6 | 4 | 6 | 9 | 9 |
| ROB | 2 | 1 | 3 | 1 | 3 | 3 |
| SET | 21 | 24 | 24 | 35 | 39 | 48 |
| SYN | 0 | 0 | 1 | 0 | 0 | 1 |
| | 118 | 77 | 145 | 103 | 172 | 211 |

Table 1: Integration of top-down/bottom-up approaches by cooperative provers: solved hard problems

The results reveal the high potential of our approach to significantly improve on single provers. Spass is only able to solve 55.9% of the problems which can be solved by cooperation, Setheo can only solve 36.5%. Competition of provers is very successful because of the very different behavior of the provers. But even a competitive prover consisting of Spass and Setheo can only solve 81.5% of the problems solvable by cooperation. Hence, cooperation is really important in order to increase the success rate. When integrating subgoal clauses into Spass its solvability rate is increased by 22.9%. In the most cases subgoal clauses take part in the search process and can help to reorder the search in a favorable manner. The use of lemmas increases Setheo's performance by 33.8%. The increase of the solvability rate of Setheo is really due to occurring resource reductions. In almost all cases where a substantial speed-up is obtained we could find a proof with a smaller resource. Then, the lemmas are used as expected, i.e. they are able to close subgoals that occur after few inferences and whose ME proof would require many inferences.

When taking a closer look at the results we can recognize the following. A prover which already shows a rather satisfactory behavior in a specific domain can often profit from others. Cooperation can entail that other hard problems can additionally be solved. However, if a prover is not suitable for a certain domain then cooperation will normally not result in a significant increase of its performance. Because of the fact that Setheo and Spass show a very different behavior in the most cases at least one prover can be improved in a certain domain.

It is interesting to find out whether certain characteristics of problems lead to a high or low performance of the cooperative system. We examine whether the characteristics "a domain contains equality" and "a domain contains non-Horn problems" influence the performance. First, we should note that these characteristics do not completely determine the performance of the cooperative system. There are gains of efficiency for all kinds of problem, regardless of the type of clauses occurring in the problems. But we can at least observe some tendencies.

Firstly, we can observe that the cooperation approach is especially well-suited for problems containing equality. The best results are obtained in the domains CAT, GRP, and SET which contain many problems with equality. When analyzing proof runs we can find two reasons for this. In such domains SPASS is able to support SETHEO because it has much stronger inferences for handling equality than SETHEO. SPASS can often derive "difficult" lemmas with few inferences, i.e. lemmas whose derivation would require many inferences by SETHEO. SETHEO can support SPASS because it is able to make transformations of the proof goal that SPASS cannot perform because of its fixed ordering used for superposition. This can increase the flexibility of the proof search performed by SPASS.

Secondly, we consider whether the fact that a domain contains mostly Horn or non-Horn problems influences the performance of the cooperation approach. Considering the domains where the cooperation approach could successfully be applied we can notice that successes could be obtained for Horn (e.g., COL) as well as non-Horn domains (e.g., SET). In the domains where no hard problems could be solved (neither sequentially nor with cooperation) often the percentage of non-Horn clauses is rather high (note that these domains do not appear in the table). The main reason for this, however, appears to be that the single provers show a weak performance in these domains. A strong relationship between the performance of the cooperative prover and the fact whether a problem is Horn or non-Horn could not really be found in the experiments.

### 5.2.2 ANALYSIS OF RUNTIMES IN SELECTED DOMAINS

Up to now we only considered the number of solved problems. In addition, it is interesting to analyze whether the use of subgoal clauses or lemmas can speed-up the proof search in general, i.e. also for problems that can be solved by single provers. Short run times are especially important if theorem provers are used within interactive prover environments. We restrict ourselves to the three domains CAT, COL, and LDA and are going to analyze the runtimes in more detail.

Table 2 presents the runtimes when tackling hard problems of the three test domains. We omitted all problems that could neither be solved by a single prover when working alone, nor by any of the cooperation variants. Column 1 of the table displays the name of the problem. Columns 2 and 3 present the runtimes of SPASS and SETHEO (on a SPARCstation-20/712) when working alone, columns 4 and 5 the runtimes of SPASS when it obtains subgoal clauses from SETHEO which are generated regarding variants 1 and 2, respectively. Note that the runtimes include the generation and selection time of subgoal clauses, and the transmission to SPASS. Column 6 displays the runtime of SETHEO if it obtains bottom-up generated lemmas from SPASS. In that case we always employed variant 2 for generating subgoal clauses. Also these runtimes include the preprocessing of SPASS and the transmission and integration of the lemmas. Column 7 gives the runtime of a competitive version of SPASS and SETHEO (minimum of the runtimes of columns 2 and 3). Finally, in column 8 we can find the runtime of our cooperative system (minimum of the runtimes of columns 5 and 6). The entry "–" means that the problem could not be solved within 1000 seconds.

| problem | SPASS | SETHEO | inference-based | adaptive | lemma | competitive | cooperative |
|---|---|---|---|---|---|---|---|
| LDA005-2 | 279s | – | 265s | 8s | – | 279s | 8s |
| LDA006-2 | 276s | – | 304s | 10s | – | 276s | 10s |
| LDA007-1 | 16s | 366s | 19s | 21s | 311s | 16s | 21s |
| LDA007-2 | – | 50s | 7s | 7s | 14s | 50s | 7s |
| LDA009-2 | – | – | – | 24s | – | – | 24s |
| LDA010-1 | – | – | – | 9s | – | – | 9s |
| LDA010-2 | – | – | – | 26s | – | – | 26s |
| LDA011-1 | 54s | – | 58s | 9s | – | 54s | 9s |
| LDA011-2 | 21s | – | 35s | 7s | – | 21s | 7s |
| CAT001-1 | – | – | – | – | 9s | – | 9s |
| CAT001-3 | 134s | 32s | 6s | 6s | 11s | 32s | 6s |
| CAT001-4 | 33s | 11s | 5s | 5s | 6s | 11s | 5s |
| CAT002-2 | – | – | – | – | 98s | – | 98s |
| CAT003-1 | – | – | – | – | 38s | – | 38s |
| CAT004-3 | – | 23s | – | 9s | 10s | 23s | 9s |
| CAT008-1 | 91s | 126s | 6s | 6s | 48s | 91s | 6s |
| CAT009-1 | – | – | 10s | 10s | – | – | 10s |
| CAT009-3 | – | – | – | 29s | 17s | – | 17s |
| CAT009-4 | 53s | – | 47s | 50s | – | 53s | 50s |
| CAT010-1 | – | – | 11s | 9s | – | – | 9s |
| CAT011-3 | 17s | – | 12s | 12s | – | 17s | 12s |
| CAT014-3 | 18s | – | 11s | 11s | – | 18s | 11s |
| CAT018-3 | – | – | – | 74s | – | – | 74s |
| COL003-2 | – | – | – | – | 494s | – | 494s |
| COL003-3 | – | 60s | – | – | 54s | 60s | 54s |
| COL003-4 | – | 19s | – | – | 35s | 19s | 35s |
| COL003-5 | – | – | – | – | 100s | – | 100s |
| COL003-7 | – | 285s | – | – | 32s | 285s | 32s |
| COL003-9 | – | 27s | – | – | 21s | 27s | 21s |
| COL034-1 | – | 60s | – | – | 70s | 60s | 70s |
| COL036-1 | – | 108s | – | – | 106s | 108s | 106s |
| COL037-1 | – | 110s | – | – | 36s | 110s | 36s |
| COL038-1 | – | 110s | – | – | 108s | 110s | 108s |
| COL041-1 | – | 39s | – | – | 38s | 39s | 38s |
| COL042-2 | – | – | – | – | 48s | – | 48s |
| COL042-3 | – | – | – | – | 81s | – | 81s |
| COL042-4 | – | – | – | – | 52s | – | 52s |
| COL057-1 | – | 12s | – | – | 8s | 12s | 8s |
| COL060-1 | 46s | – | 20s | 20s | – | 46s | 20s |
| COL061-1 | 46s | – | 17s | 16s | – | 46s | 16s |
| | 32.5% | 40.0% | 40.0% | 55.0% | 62.5% | 62.5% | 100% |

Table 2: Integration of top-down/bottom-up approaches by cooperative provers: runtimes

Since all domains contain equality the results are better than the results over the whole TPTP. Our cooperative prover can solve all listed problems, whereas SPASS is only able to solve 32.5%, SETHEO only 40.0%. A competitive prover consisting of SPASS and SETHEO can merely solve 62.5% of the problems. Not only the success rate but also the runtimes are clearly improved when using a cooperative prover. The runtimes are often decreased by substantial factors (in spite of the fact that running the two provers in parallel consumes twice as much total CPU time).

When studying the runtimes and proofs obtained by SETHEO we can observe the following. If speed-ups of SETHEO are really due to occurring resource reductions, in almost all cases a substantial speed-up is obtained. Sometimes—for instance in the COL domain—we have the situation where no resource reduction takes place but reordering effects allow finding proofs faster. In this situation, the speed-ups are low. Let us take a closer look at the runtimes of SPASS when using subgoal clauses. When considering the results of variant 1, the results show that a naive and uninformed generation of subgoal clauses usually does not entail much gain. So, only 40% of the problems can be solved using this variant. Variant 2, however, shows quite a satisfactory behavior. Hence, an intelligent generation of a subgoal clause pool really does strongly influence the efficiency.

## 6. Discussion

Integration of top-down and bottom-up provers by employing cooperation is very promising in the field of automated deduction. Due to certain strengths and weaknesses of provers following different paradigms, techniques that try to combine the strengths by cooperation can allow an improvement of the deductive system. Our approach of combining top-down and bottom-up provers by processing top-down generated subgoal clauses in a bottom-up prover achieves this combination by introducing goal-orientation into a bottom-up prover thus combining strong redundancy control mechanisms and goal-directed search. The use of bottom-up generated lemmas in a top-down prover can contribute to significantly reduce proof lengths such that proofs can be found with smaller resources.

Related approaches for supporting top-down by bottom-up inference also mainly aimed at employing bottom-up created lemmas in a top-down prover. Similar to our method by Schumann (1994) and Fuchs (1998a, 1999) lemmas are created in a preprocessing phase and the input clauses are augmented by these formulas. The main difference of these approaches and our approach is the kind of the used lemmas. There, the ME inference mechanism is used in order to generate lemmas. This has the advantage that in some cases—in contrast to our technique—proof length or resource reductions are guaranteed. However, the lemma mechanisms used by Fuchs (1998b, 1998a, 1999) generate quite "easy" lemmas. Hence, their potential w.r.t. the size of the resource reduction is limited.

Other approaches try to dynamically create unit lemmas during the proof run of the ME prover (Astrachan & Stickel, 1992; Iwanuma, 1997; Astrachan & Loveland, 1997).

After each successful solution of a subgoal a lemma might be generated and added to the input clauses. The aim of this kind of lemma generation is to produce lemmas that are able to reduce the search amount by eliminating repeated sub-deductions. One criticism regarding this kind of lemma generation is the fact that it is unclear whether or not useful lemmas can be generated. There is no guarantee that lemmas can be produced during the proof run which can contribute to a proof, i.e. which can be "re-used". Furthermore, as already mentioned, the generated lemmas are usually not as general as possible due to instantiations coming from the solutions of subgoals previously solved. This can reduce the applicability of a lemma although the "generalized" proof could be re-used for refuting the input clauses. Thus, although some hard problems could only be solved with such lemma techniques (see Astrachan & Loveland, 1997), no stable success has been reported over a large set of problems. The main disadvantages of all approaches which only aim at supporting top-down provers originate from the fact that in some domains, especially if equality is involved, superposition-based provers clearly outperform ME provers. Thus, in such domains it may be more sensible to develop techniques in order to support the more powerful bottom-up prover than the weaker top-down prover.

In order to improve bottom-up proof search by using top-down performed inferences the following approaches have been employed. Firstly, again one prover (the bottom-up prover) is assisted by clauses derived from another prover (the top-down prover). The approach from Sutcliffe (1992) uses lemmas generated by a guided linear deduction system (and not subgoal clauses) in order to support resolution-based provers. Due to the lack of goal orientation (as described in Section 2.2) this method could not yield convincing results in practice. Secondly, there are approaches to make bottom-up provers more goal-directed by forcing them to work only with some relevant clauses which are detected by top-down calculations. The methods described by Bancilhon et al. (1986), Stickel (1994), and Hasegawa et al. (1997) transform a set of clauses into another clause set which is then evaluated in a bottom-up manner. The specific transformation provides a combination of top-down and bottom-up processing and prunes the bottom-up evaluation to relevant clauses (which bear a connection to a proof goal). Thus, obviously the bottom-up proof search becomes more goal oriented. Also the method described by Loveland et al. (1995) provides a relevancy testing for bottom-up calculations. Based on top-down proof attempts the relevancy of a clause is dynamically determined during the bottom-up calculation. In contrast to our method in these approaches the bottom-up prover has to tackle the whole proof task. Our approach for using top-down generated subgoal clauses in a bottom-up prover does not provide a relevancy testing of bottom-up inference but supports the bottom-up inference process by simplifying the original goal. Thus, the proof length may be shortened. Furthermore, parts of the search space of the bottom-up prover, which contain relevant clauses but may be difficult to enumerate, can be traversed by a single inference which provides large search reductions.